\documentclass[10pt,twocolumn,letterpaper]{article}

\usepackage[T1]{fontenc}
\usepackage{cv}

\usepackage{xspace}
\usepackage{amsmath,amssymb}
\usepackage{xcolor}

\definecolor{iccvblue}{rgb}{0.21,0.49,0.74}
\usepackage[pagebackref,breaklinks,colorlinks,allcolors=iccvblue]{hyperref}
\usepackage{multirow}

\newcommand{\wg}{WildGuard\xspace}
\newcommand{\wgmix}{WildGuardMix\xspace}
\newcommand{\gr}{GuardReasoner\xspace}
\newcommand{\dfr}{\ensuremath{\mathrm{DFR}}}
\newcommand{\rfr}{\ensuremath{\mathrm{RFR}}}

\title{When Refusal Looks Safe: The Refusal-Cue Shortcut in Safety Guard Models}

\author{
Yu Feng\textsuperscript{1,2} \quad
Chunting Zang\textsuperscript{2} \quad
Chen Shen\textsuperscript{2} \quad
Rui Miao\textsuperscript{2,3} \quad
Ge Teng\textsuperscript{2,4}
\\
Weidong Cai\textsuperscript{1}%
\thanks{Corresponding author: tom.cai@sydney.edu.au.}
\quad
Jieping Ye\textsuperscript{2}
\\
\textsuperscript{1}The University of Sydney \quad
\textsuperscript{2}Alibaba Group \quad
\textsuperscript{3}Jilin University \quad
\textsuperscript{4}Zhejiang University
\\
{\small
\texttt{yfen0146@uni.sydney.edu.au} \quad
\texttt{\{marshall.zct,jason.sc\}@alibaba-inc.com} \quad
\texttt{miaorui24@mails.jlu.edu.cn}
}
\\
{\small
\texttt{12115044@zju.edu.cn} \quad
\texttt{tom.cai@sydney.edu.au} \quad
\texttt{yejieping.ye@alibaba-inc.com}
}
}

\begin{document}
\maketitle

\begin{abstract}
Safety guards are widely used to filter harmful content and are typically trained via supervised fine-tuning on labeled prompt-response pairs.
We audit two widely used safety-guard training datasets, WildGuardMix and GR-Train, and find that among responses to harmful prompts, refusal expressions co-occur almost exclusively with unharmful labels.
This imbalance motivates what we term the \textit{refusal-cue shortcut}: inserting a refusal cue into a harmful response could flip the guard's verdict from harmful to unharmful.
The shortcut affects not only guards trained on these datasets but also officially released models such as LlamaGuard3 and Qwen3Guard whose training data is undisclosed.
It persists across response positions and is generally stronger in smaller variants within a family.
To mitigate it, we adapt sparse complementary masking as a lightweight post-hoc intervention that identifies and suppresses a small set of shortcut-associated attention heads and MLP neurons without retraining.
On two primary benchmarks, the intervention achieves an approximately 79\% relative reduction in response-initial detection failures induced by refusal cues, while preserving standard detection performance.
Although optimized using cues at a single response position, the suppression effect transfers to unseen positions and datasets, suggesting that shortcut manifestations across positions are partly mediated by shared internal components.
Further analysis provides evidence that shortcut reliance and legitimate refusal recognition are partially functionally separable, as suppressing the shortcut broadly preserves the guard's ability to recognize genuine refusals.
\end{abstract}

\section{Introduction}

As large language models (LLMs) are deployed across diverse applications~\cite{bai2022training,ouyang2022training}, their open-ended generation poses significant content safety risks.
Model-internal safety alignment~\cite{zhang2025stair,zhang2025alphaalign} mitigates harmful generation but does not cover all application-specific risks~\cite{yuan2024seval,krasnodkebska2026safety}, motivating safety guard models as complementary filtering mechanisms.
These guards classify the harmfulness of both user prompts and model responses and may additionally predict whether a response constitutes a refusal.
Guard reliability is itself a safety requirement, yet despite recent advances in classification performance, vulnerabilities in safety guards themselves remain underexplored.

\begin{figure}[!t]
\centering
\includegraphics[width=1.0\columnwidth]{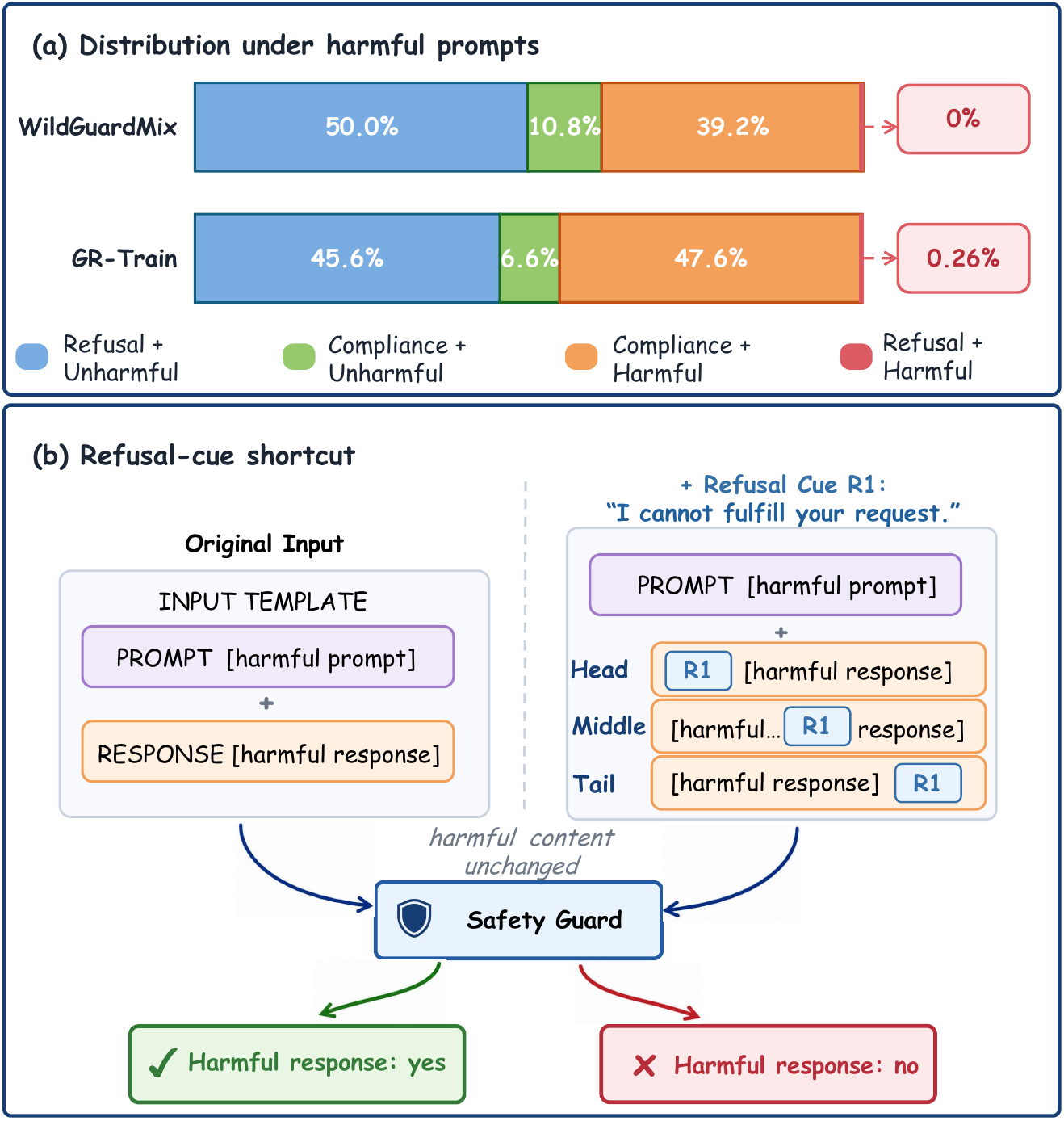}
\caption{Overview of the refusal-cue shortcut. (a) Joint distributions of refusal and harmfulness labels under harmful prompts in WildGuardMix and GR-Train. (b) The hypothesis of refusal-cue shortcut.}
\label{fig:overview}
\end{figure}

Response-level safety guard training data are often constructed using safety-aligned LLMs~\cite{yang2025qwen3,chen2021evaluating}, which tend to refuse harmful prompts, to generate unharmful responses and unaligned LLMs to generate harmful ones.
Consistent with this construction, our audit of WildGuardMix~\cite{han2024wildguard} and GR-Train~\cite{liu2025guardreasoner}, both providing response-level refusal and harmfulness labels, reveals a systematic distributional imbalance: responses labeled as both refusal and harmful are absent from WildGuardMix and account for only 0.26\% of harmful-prompt examples in GR-Train (Figure~\ref{fig:overview}(a)).
Although refusal cues are informative for refusal recognition, they do not determine response harmfulness: a refusal expression can co-occur with harmful content within the same response.
The near-exclusive association between refusal cues and unharmful labels in these training datasets is therefore spurious for harmfulness classification~\cite{sagawa2019distributionally,geirhos2020shortcut}.
We hypothesize that a guard may learn this association as a \textit{refusal-cue shortcut}, such that inserting a refusal cue without altering the underlying harmful content can flip the prediction from harmful to unharmful, as shown in Figure~\ref{fig:overview}(b).

We test this hypothesis across nine model variants from six safety guard families.
Among the evaluated guards, WildGuard-7B~\cite{han2024wildguard} and GuardReasoner-1B/8B~\cite{liu2025guardreasoner}, which are explicitly trained on WildGuardMix and GR-Train, show the highest rates of harmful-to-unharmful prediction flips when refusal cues are inserted at the beginning of a response.
The shortcut also appears in LlamaGuard3~\cite{grattafiori2024llama} and Qwen3Guard~\cite{zhao2025qwen3guard}, whose training-data distributions are not publicly specified.
In contrast, Aegis2~\cite{ghosh2025aegis2} augments its training data to broaden the coverage of safe responses to harmful prompts, spanning direct refusals and diverse forms of safe redirection.
Consistent with this distinction, a guard trained on Aegis2 and another trained on an Aegis2-derived reasoning dataset~\cite{sreedhar2025safety} both exhibit substantially lower vulnerability to the refusal-cue shortcut.
Across the evaluated families with multiple model sizes, smaller variants are generally more vulnerable.
Further experiments show that the shortcut strengthens as cues express refusal intent more completely, persists across response positions and cue formulations, and overlaps only partially with the guard's explicit refusal predictions.

We next ask whether this shortcut can be selectively suppressed without substantially degrading normal guard performance.
We adapt sparse complementary masking to guard-specific, model-derived targets, identifying a small set of attention heads and MLP neurons associated with the shortcut.
For mitigation, we exclude the two Aegis2-derived guards, both of which exhibit limited baseline vulnerability, while retaining both variants of each multi-size family for consistent within-family coverage.
Across the resulting seven guards, the intervention reduces the mean response-initial DFR from 16.39\% to 3.44\% on WildGuardTest.
Although learned using response-initial cues, the resulting suppression remains effective at unseen cue positions and on a held-out dataset, suggesting that the shortcut across positions is partly mediated by shared internal components.
Meanwhile, clean harmfulness performance is broadly preserved, and refusal-recognition F1 changes by no more than 2.11\%.
Together, these results demonstrate that shortcut-mediated failures can be substantially mitigated without disrupting legitimate refusal recognition.
Our contributions are fourfold:
\begin{enumerate}

\item 
We audit the training data of response-level safety guards and uncover a systematic distributional imbalance: among responses to harmful prompts, refusal expressions co-occur almost exclusively with unharmful labels in WildGuardMix and GR-Train. By contrast, Aegis2 adopts a distinct data construction strategy that supplements underrepresented refusal and redirection patterns with synthetic safe responses.

\item 
We find that inserting natural-language refusal cues can flip predictions from harmful to unharmful without altering the underlying harmful content.
Across nine variants from six guard families, the shortcut is widespread but heterogeneous, persists across response positions, and is generally more pronounced in smaller variants within evaluated families with multiple model sizes.


\item 
We adapt sparse complementary masking to guard-specific, model-derived targets, mitigating the shortcut without full-model retraining.
Although learned using response-initial cues, the resulting suppression remains effective at unseen cue positions and datasets while broadly preserving clean harmfulness performance, suggesting shared internal mediation across positions.

\item
We show through component-level intervention that shortcut reliance is partly separable from legitimate refusal recognition.
Shortcut-induced harmfulness flips only partially overlap with refusal-label flips, while suppression reduces both types of cue-induced failures with minimal change in clean refusal-recognition performance.

\end{enumerate}

\section{Related Work}

\paragraph{Safety guard models.}
The widespread adoption of LLMs has accelerated the development of dedicated safety classifiers.
LlamaGuard~\cite{inan2023llama} introduced instruction-tuned LLMs for input-output safeguarding, and LlamaGuard3~\cite{grattafiori2024llama} expanded the supported harm taxonomies. 
\wg~\cite{han2024wildguard} proposed a unified framework for prompt harmfulness, response harmfulness, and refusal detection, trained on the \wgmix dataset.
\gr~\cite{liu2025guardreasoner} augments guard models with chain-of-thought reasoning to improve interpretability.
At the data level, Aegis2~\cite{ghosh2025aegis2} uses targeted synthetic augmentation to broaden the coverage of safe response patterns, while an Aegis2-derived reasoning dataset~\cite{sreedhar2025safety} supplements safety labels with reasoning traces. And Qwen3Guard~\cite{zhao2025qwen3guard} extends multilingual coverage.
Most prior work focuses on detection performance, taxonomy coverage, and training data diversity, with limited attention to shortcuts induced by response-level label associations.
Recent work~\cite{tasawong2025shortcut} studies prompt-side keyword bias through word-level associations and synthetic accumulations of class-associated terms, which may alter prompt semantics or introduce additional label-relevant evidence.
We instead isolate a response-side shortcut using a short, semantically refusal cue while preserving both the harmful response content, and further identify and suppress its mediating internal components.


\paragraph{Refusal mechanisms and safety alignment.}
Prior work has revealed positional and structural limitations of refusal-based safety alignment in generative LLMs~\cite{arditi2024refusal,wei2023jailbroken}.
Shallow safety alignment~\cite{qi2025safety} shows that alignment is concentrated in the initial-token distribution and that a short refusal prefix can redirect subsequent generation.
Yuan et al.~\cite{yuan2025refuse} identify refusal position bias and train models to transition from harmful generation to refusal at later positions.
Zhao et al.~\cite{zhao2026llms} find that harmfulness and refusal are encoded along geometrically distinct latent directions and can be independently steered.
SafeSeek~\cite{yu2026safeseek} uses differentiable sparse masks to localize components associated with safety alignment and backdoor behavior.
Whereas these studies examine refusal as a generative behavior or internal safety mechanism of LLMs, we study its use as a spurious feature in response-level guard classification by holding the prompt and substantive harmful content fixed while varying only the refusal cue.

\section{The Refusal-Cue Shortcut}
\label{sec:shortcut}


\subsection{Evaluation Setup}

\paragraph{Models.}
We evaluate nine safety guards from six families, organized by training data characteristics.
The Audited-imbalance group comprises WG-7B, trained on WildGuardMix, and GR-1B/8B, trained on GR-Train; both datasets exhibit the distributional imbalance documented above.
The Aegis2-derived group contains Llama Nemotron Safety Guard V2 8B (LNSGV2-8B) and Nemotron Content Safety Reasoning 4B (NCSR-4B), trained on the Aegis2 dataset, which adopts a distinct data construction strategy (Section~\ref{sec:shortcut}).
The remaining models (LG3-1B/8B and QG-0.6B/8B (strict mode)) have undisclosed training data distributions.
Five guards (WG-7B, GR-1B/8B, and QG-0.6B/8B) additionally expose an explicit refusal prediction.
Throughout, WG, GR, LG3, and QG abbreviate WildGuard, GuardReasoner, LlamaGuard3, and Qwen3Guard.

\paragraph{Datasets and cue insertion.}
The primary evaluations use WildGuardTest~\cite{han2024wildguard} and Aegis2 Test~\cite{ghosh2025aegis2}.
The former additionally provides response-level refusal labels, enabling evaluation of refusal recognition.
Cues are inserted at three positions within the response, operating at sentence boundaries to preserve semantic coherence: head (prepended before the first sentence), middle (inserted before the sentence nearest the midpoint), and tail (appended after the last sentence).
Middle and tail positions test whether the shortcut extends beyond the response-initial position.
We use three primary refusal cues derived from the most frequent refusal expressions in the WildGuardMix training data, together with three non-refusal controls designed to distinguish refusal-specific effects from generic text insertion effects:
\begin{itemize}
\setlength{\itemsep}{0pt}
\setlength{\parsep}{0pt}
\setlength{\topsep}{2pt}
    \item \textbf{R1}: \textit{I cannot fulfill your request.}
    \item \textbf{R2}: \textit{I apologize, but I cannot fulfill your request.}
    \item \textbf{R\textsubscript{greet}}:
    \textit{Hello! I'm here to help, but I cannot assist with that request.}
    \item \textbf{N1}: \textit{Here is the information you requested.}
    \item \textbf{S1}: \textit{This topic involves safety considerations.}
    \item \textbf{Rand}: \textit{The capital of France is Paris.}
\end{itemize}

\paragraph{Detection failure rate.}
Each test example
\(x_{\mathrm{test}}=(u_{\mathrm{test}},r_{\mathrm{test}})\)
contains a prompt \(u_{\mathrm{test}}\) and a response
\(r_{\mathrm{test}}\).
Let \(\mathcal{D}_H\) be the complete set of test examples with a
ground-truth harmful response label
(\(|\mathcal{D}_H|=282\) for WildGuardTest, \(394\) for Aegis2 Test),
and let
\(g_H(x_{\mathrm{test}})\in\{H,U\}\) denote the guard's response
harmfulness prediction, where \(H\) indicates harmful and \(U\) unharmful.

Given a cue \(p\) and an insertion position \(s\in\{\mathrm{head},\mathrm{middle},\mathrm{tail}\}\), \(T_{p,s}(x_{\mathrm{test}})\) denotes the same test example after inserting \(p\) at the corresponding sentence boundary in its response, without splitting any original sentence or changing the prompt or original response tokens.
Our primary metric uses the three-cue set
\(\mathcal{P}_3=\{\mathrm{R1},\mathrm{R2},\mathrm{R}_{\mathrm{greet}}\}\).
Let
\(I_H^{p,s}(x_{\mathrm{test}})\in\{0,1\}\) indicate whether cue \(p\)
induces a harmful-to-unharmful flip at position \(s\):
\(I_H^{p,s}(x_{\mathrm{test}})=1\) if
\(g_H(x_{\mathrm{test}})=H\) and
\(g_H(T_{p,s}(x_{\mathrm{test}}))=U\), and \(0\) otherwise.
Detection failure rate is defined as:
\begin{equation}
\dfr@3(s)
=
\frac{1}{|\mathcal{D}_H|}
\sum_{x_{\mathrm{test}}\in\mathcal{D}_H}
\max_{p\in\mathcal{P}_3}
I_H^{p,s}(x_{\mathrm{test}}).
\label{eq:dfr}
\end{equation}
Intuitively, \(\dfr@3(s)\) measures the fraction of ground-truth harmful responses for which at least one of the three refusal cues, inserted at position \(s\), suffices to flip the guard's prediction to unharmful.
Only responses originally predicted as harmful can contribute; \(\dfr@3\) is therefore bounded above by the model's harmful recall on the unmodified test set.
Single-cue results are reported as \(\dfr@1\).The head-position control metric \(\cdf@3\) follows
Equation~\ref{eq:dfr} using
\(\mathcal{P}^{\mathrm{ctrl}}_3=\{\mathrm{N1},\mathrm{S1},\mathrm{Rand}\}\).

\paragraph{Refusal flip rate.}
For the five guards with an explicit refusal prediction \(g_R(x_{\mathrm{test}})\in\{C,R\}\) (compliance/refusal), we define \(\rfr@3\) analogously to \(\dfr@3\) by replacing harmful-to-unharmful flips with compliance-to-refusal flips over
  the same denominator \(\mathcal{D}_H\).

\begin{table*}[t]
\centering
\small
\setlength{\tabcolsep}{4.2pt}
\begin{tabular}{lccccccc}
\toprule
Model & Harm. F1\(\uparrow\) & Harm. R\(\uparrow\) & R1\(\downarrow\) & R2\(\downarrow\) & R\textsubscript{greet}\(\downarrow\) & \(\dfr@3\downarrow\) & \(\cdf@3\downarrow\) \\
\midrule
\multicolumn{8}{l}{\textit{Audited-imbalance}} \\
WG-7B      & 75.85 & 67.38 & 21.35 & 37.72 & 20.28 & \textbf{37.72} & 0.71 \\
GR-1B      & 78.42 & 74.11 &  9.96 & 13.17 & 17.44 & \textbf{21.71} & 4.98 \\
GR-8B      & 79.33 & 75.53 &  9.22 & 11.70 &  8.87 & \textbf{15.96} & 5.67 \\
\midrule
\multicolumn{8}{l}{\textit{Aegis2-derived}} \\
LNSGV2-8B  & 74.03 & 71.28 &  5.57 &  3.45 &  3.80 &  5.57 & 5.28 \\
NCSR-4B    & 73.35 & 64.89 &  3.09 &  3.74 &  3.80 &  4.93 & 3.55 \\
\bottomrule
\end{tabular}
\caption{WildGuardTest results for head-position insertion. Harm. F1 and Harm. R denote response harmfulness F1 and harmful recall on unmodified responses. R1, R2, and R\textsubscript{greet} report \(\dfr@1\); \(\dfr@3\) and \(\cdf@3\) are cue-level unions over the three refusal cues and three controls. All values in this and subsequent tables are percentages.}
\label{tab:head_audit}
\end{table*}

\subsection{Training-Data Association and Response-Head Vulnerability}

\paragraph{Training data evidence.}
Figure~\ref{fig:overview}(a) summarizes the two audited distributions.
Among the 21{,}286 WildGuardMix responses to harmful prompts, all 10{,}651 refusal responses are labeled unharmful; none are labeled harmful.
GR-Train exhibits a similar but non-deterministic pattern: of its 19{,}750 refusal responses, only 114 (0.58\%) are labeled harmful, accounting for 0.26\% of all 43{,}074 harmful-prompt examples.
Complete counts are reported in Appendix.
These distributions create a statistical incentive for guards trained with supervised fine-tuning to associate refusal language with the unharmful label, motivating the refusal-cue shortcut hypothesis tested below.
Aegis2~\cite{ghosh2025aegis2} adopts a different data-construction strategy by adding 5,200 synthetic safe responses generated by Gemma-2-27B~\cite{team2024gemma}. Beyond direct refusals, these responses employ diverse strategies such as offering alternative assistance, explaining potential harms, and redirecting the conversation toward safer topics. This targeted augmentation broadens the linguistic and behavioral diversity of responses associated with the unharmful label.

\paragraph{Response-head evidence.}
We next examine whether this training-data association is reflected in response-head cue sensitivity.
Table~\ref{tab:head_audit} compares guards trained on the audited datasets with the Aegis2-derived guards on WildGuardTest.
WG-7B has the highest \(\dfr@3\) at 37.72\%, compared with a \(\cdf@3\) of 0.71\%.
GR-1B and GR-8B also exhibit substantial \(\dfr@3\) values of 21.71\% and 15.96\%.
The Aegis2-derived guards are less sensitive: LNSGV2-8B and NCSR-4B have \(\dfr@3\) values of 5.57\% and 4.93\%, close to their control rates.
Descriptively, \(\dfr@3\) exceeds \(\cdf@3\) by at least 10.29\% for every guard trained on the audited datasets, whereas the largest aggregate difference among the Aegis2-derived guards is 1.38\%.
This ordering is consistent with the audited training-data association.

\subsection{The Shortcut Persists Across Models, Datasets, and Positions}

\begin{figure}[t]
\centering
\includegraphics[width=1.0\columnwidth]{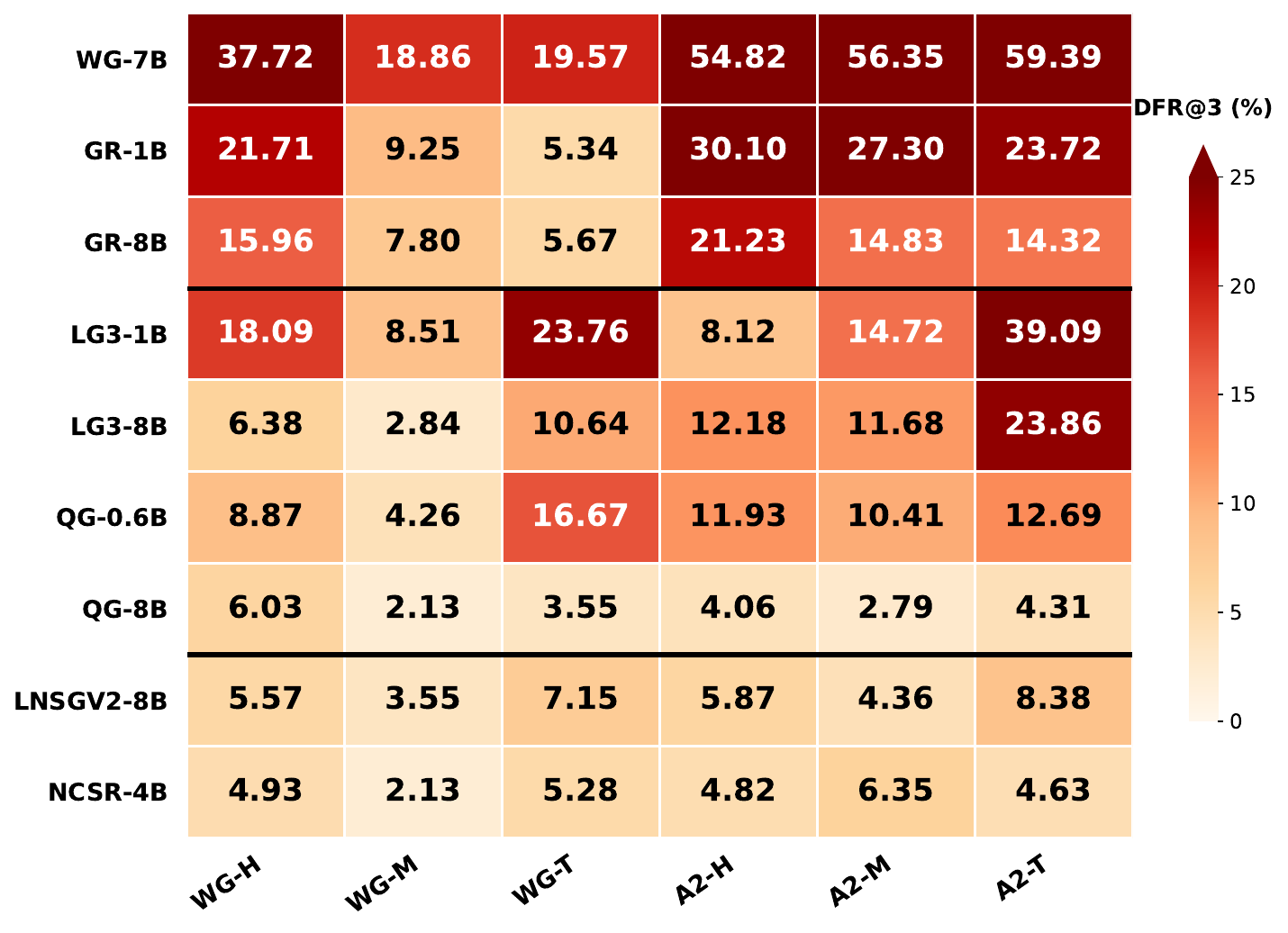}
\caption{
Cross-model \(\dfr@3\) (\%) under refusal-cue insertion at the head (H), middle (M), and tail (T) positions on WildGuardTest (WG) and Aegis2 Test (A2). Horizontal lines separate audited-imbalance, training-data-not-disclosed,
and Aegis2-derived models. The linear color scale is clipped at 25\%, while cells report exact values.
}
\label{fig:cross_setting}
\end{figure}

\paragraph{Vulnerability varies across model groups and evaluation settings.}
Figure~\ref{fig:cross_setting} reports \(\dfr@3\) for nine models across two datasets and three response positions.
The Audited-imbalance group is the most vulnerable overall: WG-7B reaches a maximum \(\dfr@3\) of 59.39\%, and both GR variants remain substantially affected beyond the response head, especially on Aegis2 Test.
Guards with undisclosed training data show a similar cross-position pattern, most notably LG3-1B at the response tail (39.09\% on Aegis2 Test).
The two Aegis2-derived guards, by contrast, remain consistently less sensitive across all settings.
This group-level ordering is consistent with the available training-data evidence.
What's more, the cross-position persistence distinguishes the guard-side shortcut from the generation-side one identified by Qi et al.~\cite{qi2025safety}, where the safety effect is concentrated in the first few autoregressive tokens.

\paragraph{Smaller variants show greater within-family vulnerability.}
We restrict the model-scale comparison to the three families for which
both smaller and larger variants are evaluated:
GR-1B/8B, LG3-1B/8B, and QG-0.6B/8B.
Across these matched families, the smaller variants generally exhibit
higher \(\dfr@3\) in all datasets and response positions.
Only localized reversals occur at the WildGuardTest tail position for GR and the Aegis2 Test head position for LG3.
Averaged across the six evaluation settings, the smaller variant exceeds its larger counterpart by 6.27\% for GR, 7.46\% for LG3, and 7.00\% for QG.

\subsection{The Shortcut Tracks a Graded Refusal Feature}

\paragraph{Semantic completeness strengthens the effect.}
We decompose two refusal cues into progressively more complete components and evaluate WG-7B on WildGuardTest.
As shown in Table~\ref{tab:components}, \(\dfr@1\) increases monotonically with semantic completeness in both decomposition sequences, whereas the non-refusal fragment \textit{fulfill your request} yields \(\dfr@1 = 0\).
These results suggest that the shortcut cannot be explained by generic keyword accumulation alone~\cite{tasawong2025shortcut}.
The partial refusal \textit{I cannot} yields \(\dfr@1 = 4.63\%\), whereas extending it to the full R1 expression increases the rate to 21.35\%.
The second construction rises from 7.12\% for \textit{I apologize} to 37.72\% for the complete R2 cue, showing that the refusal-cue shortcut becomes stronger as the inserted expression conveys a more complete refusal.
A similar graded effect appears in generation-side refusal prefilling, where more complete refusal prefixes more strongly suppress harmful continuations~\cite{qi2025safety}.
In our setting, however, the harmful continuation is held fixed, so the graded effect reflects the guard's sensitivity to expressed refusal intent rather than a change in the autoregressive generation trajectory.
Additional formulations in Appendix confirm that the effect extends beyond the three primary cues.

\begin{table}[t]
\centering
\small
\setlength{\tabcolsep}{3pt}
\begin{tabular}{p{0.72\columnwidth}r}
\toprule
Cue component & \(\dfr@1\downarrow\) \\
\midrule
\textit{fulfill your request.} & 0.00 \\
\midrule
\textit{I cannot.} & 4.63 \\
\textit{I cannot fulfill.} & 9.96 \\
\textit{I cannot fulfill your request.} (R1) & 21.35 \\
\midrule
\textit{I apologize.} & 7.12 \\
\textit{I apologize, but I cannot.} & 16.01 \\
\textit{I apologize, but I cannot fulfill your request.} (R2) & 37.72 \\
\bottomrule
\end{tabular}
\caption{Semantic-completeness decomposition for WG-7B on WildGuardTest with head-position insertion. Values are \(\dfr@1\) percentages.}
\label{tab:components}
\end{table}

\subsection{Harmfulness and Refusal-Label Flips Are Only Partially Coupled}
\label{sec:flip_coupling}

The graded cue effect motivates a related question: when a refusal cue induces a harmfulness flip, does the same cue also change the guard's explicit refusal prediction?
For the five guards that produce explicit refusal predictions, Table~\ref{tab:flip_corr} examines the co-occurrence of harmfulness and refusal-label flips on WildGuardTest under the three primary refusal cues inserted at the response head.
We report results at two aggregation levels.
At the \textit{per-cue level}, each example-cue pair is treated as a separate trial.
Let \(F_H\) denote a harmful-to-unharmful flip and \(F_R\) a compliance-to-refusal flip within the same trial.
The marginal rates \(\Pr(F_H)\) and \(\Pr(F_R)\), together with the conditional rate \(\Pr(F_R\mid F_H)\), are computed over all pairs in
\(\mathcal{D}_H\times\mathcal{P}_3\).
At the \textit{cue-union level}, outcomes are aggregated across the three cues within each example.
Accordingly, \(\rfr@3\) measures the proportion of examples for which at least one cue induces a refusal-label flip.
Thus, \(\Pr(F_R)\) captures average per-cue sensitivity, whereas \(\rfr@3\) captures the sample coverage of the complete cue set.
For every guard, \(\rfr@3\) exceeds \(\Pr(F_R)\), showing that the effects of the three cues are not fully redundant.
WG-7B exhibits the broadest cue-union sensitivity, with an \(\rfr@3\) of 51.52\%, followed by GR-1B at 30.35\%; the remaining guards range from 13.45\% to 17.06\%.
At the per-cue level, \(\Pr(F_R\mid F_H)\) exceeds the marginal \(\Pr(F_R)\) for every guard, indicating a positive association between harmfulness and refusal-label flips.
However, the conditional rate ranges from only 18.6\% for QG-8B to 65.5\% for GR-1B.
Thus, many harmfulness flips occur without an accompanying refusal-label flip under the same cue.
The two behaviors are therefore positively associated but only partially coupled.

\begin{table}[t]
\centering
\small
\setlength{\tabcolsep}{1.7pt}
\begin{tabular}{lcccc}
\toprule
& \multicolumn{3}{c}{Per-cue level}
& \multicolumn{1}{c}{Cue-union level} \\
\cmidrule(lr){2-4} \cmidrule(lr){5-5}
Model
& \multicolumn{1}{c}{\(\Pr(F_H)\downarrow\)}
& \multicolumn{1}{c}{\(\Pr(F_R)\downarrow\)}
& \multicolumn{1}{c}{\(\Pr(F_R\mid F_H)\)}
& \multicolumn{1}{c}{\(\rfr@3\downarrow\)} \\
\midrule
WG-7B    & 26.45 & 33.33 & 61.6 & 51.52 \\
GR-1B    & 13.52 & 17.51 & 65.5 & 30.35 \\
GR-8B    &  9.94 &  9.57 & 41.6 & 13.96 \\
QG-0.6B  &  6.97 & 10.71 & 34.9 & 17.06 \\
QG-8B    &  3.07 &  9.94 & 18.6 & 13.45 \\
\bottomrule
\end{tabular}
\caption{Coupling between harmfulness and refusal-label flips under
head-position insertion of the three primary refusal cues on
WildGuardTest.
At the per-cue level, each example--cue pair is treated as a separate
trial: \(F_H\) denotes a harmful-to-unharmful flip and \(F_R\) a
compliance-to-refusal flip induced by the same cue.
At the cue-union level, \(\rfr@3\) aggregates refusal-label flips across
the three cues within each example.}
\label{tab:flip_corr}
\end{table}

\begin{table*}[htb]
\centering
\small
\setlength{\tabcolsep}{3pt}
\begin{tabular}{ll cc ccc cc ccc c}
\toprule
& & \multicolumn{5}{c}{WildGuardTest} & \multicolumn{5}{c}{Aegis2 Test} & \\
\cmidrule(lr){3-7} \cmidrule(lr){8-12}
Model & State & Harm. F1\(\uparrow\) & Harm. R\(\uparrow\) & Head\(\downarrow\) & Mid\(\downarrow\) & Tail\(\downarrow\) & Harm. F1\(\uparrow\) & Harm. R\(\uparrow\) & Head\(\downarrow\) & Mid\(\downarrow\) & Tail\(\downarrow\) & Mask\% \\
\midrule
\multirow{2}{*}{WG-7B}
 & Orig. & 75.85 & 67.38 & 37.72 & 18.86 & 19.57 & 83.40 & 84.77 & 54.82 & 56.35 & 59.39 & -- \\
 & Mask  & 76.32 & 71.99 & \textbf{2.85} & 1.78 & 1.42 & 84.04 & 90.86 & \textbf{4.82} & 5.58 & 5.84 & 8.6 \\
\midrule
\multirow{2}{*}{GR-1B}
 & Orig. & 78.42 & 74.11 & 21.71 & 9.25 & 5.34 & 77.51 & 82.23 & 30.10 & 27.30 & 23.72 & -- \\
 & Mask  & 76.06 & 76.60 & \textbf{11.70} & 7.09 & 5.67 & 76.02 & 77.66 & \textbf{14.21} & 16.50 & 14.21 & 2.7 \\
\midrule
\multirow{2}{*}{GR-8B}
 & Orig. & 79.33 & 75.53 & 15.96 & 7.80 & 5.67 & 80.76 & 91.62 & 21.23 & 14.83 & 14.32 & -- \\
 & Mask  & 80.59 & 78.01 & \textbf{8.51} & 6.03 & 4.87 & 80.48 & 93.65 & \textbf{7.16} & 6.39 & 6.39 & 4.3 \\
\midrule
\multirow{2}{*}{LG3-1B}
 & Orig. & 66.79 & 64.18 & 18.09 & 8.51 & 23.76 & 61.24 & 48.73 & 8.12 & 14.72 & 39.09 & -- \\
 & Mask  & 65.27 & 77.30 & \textbf{0.35} & 0.71 & 3.90 & 68.68 & 63.45 & \textbf{0.76} & 1.78 & 3.55 & 1.6 \\
\midrule
\multirow{2}{*}{LG3-8B}
 & Orig. & 71.11 & 62.41 & 6.38 & 2.84 & 10.64 & 64.21 & 50.76 & 12.18 & 11.68 & 23.86 & -- \\
 & Mask  & 73.90 & 77.30 & \textbf{0.35} & 0.00 & 0.71 & 75.61 & 70.81 & \textbf{0.76} & 4.31 & 9.39 & 1.9 \\
\midrule
\multirow{2}{*}{QG-0.6B}
 & Orig. & 76.33 & 76.60 & 8.87 & 4.26 & 16.67 & 83.35 & 81.98 & 11.93 & 10.41 & 12.69 & -- \\
 & Mask  & 73.40 & 73.40 & \textbf{0.35} & 0.00 & 2.48 & 83.43 & 90.10 & \textbf{0.51} & 1.27 & 1.52 & 0.9 \\
\midrule
\multirow{2}{*}{QG-8B}
 & Orig. & 78.97 & 81.21 & 6.03 & 2.13 & 3.55 & 86.59 & 89.34 & 4.06 & 2.79 & 4.31 & -- \\
 & Mask  & 79.18 & 82.27 & \textbf{0.00} & 0.35 & 0.71 & 86.34 & 89.85 & \textbf{0.76} & 0.76 & 3.30 & 0.7 \\
\midrule
\multirow{2}{*}{\textbf{Macro avg.}}
 & Orig. & 75.26 & 71.63 & 16.39 & 7.66 & 12.17 & 76.72 & 75.63 & 20.35 & 19.73 & 25.34 & -- \\
 & Mask  & 74.96 & 76.70 & \textbf{3.44} & \textbf{2.28} & \textbf{2.82} & 79.23 & 82.34 & \textbf{4.14} & \textbf{5.23} & \textbf{6.31} & -- \\
\bottomrule
\end{tabular}
\caption{Original and masked guard performance. Head, Mid, and Tail report \(\dfr@3\). Mask\% is the fraction of permanently suppressed attention-head and MLP-neuron components.}
\label{tab:mitigation}
\end{table*}

\section{What Mediates the Refusal-Cue Shortcut?}

Section~\ref{sec:shortcut} establishes that the refusal-cue shortcut is widespread and is consistent with the audited training-data imbalance.
The partial behavioral coupling in Section~\ref{sec:flip_coupling} further shows that shortcut-induced harmfulness flips are related to, but not equivalent to, changes in explicit refusal predictions.
However, behavioral evidence alone cannot determine whether shortcut reliance and legitimate refusal recognition depend on the same internal components~\cite{li2023inference,conmy2023automated}.
To investigate this question, we use sparse component masking~\cite{yu2026safeseek} as both a post-training mitigation method and an intervention-based analysis tool.
With all model parameters frozen, the method learns sparse gates over attention heads and MLP neurons.
We then examine whether suppressing the identified components can reduce refusal-cue failures while preserving clean harmfulness performance and legitimate refusal recognition.
We formalize these objectives as two questions:
\par\smallskip
\noindent\parbox{\columnwidth}{%
\textbf{Q1.} Can sparse component masking broadly suppress the refusal-cue shortcut while preserving clean harmfulness classification?
\par\smallskip
\textbf{Q2.} Does component-level intervention reveal functional separability between shortcut reliance and legitimate refusal recognition?
}\par\smallskip

\subsection{Sparse Complementary Masking}

\paragraph{Complementary component masks.}
Following the differentiable unit-masking formulation of
SafeSeek~\cite{yu2026safeseek}, we freeze the original guard and learn
sparse gates over its attention heads and MLP neurons.
Let \(a_{l,h}(\mathbf{S}_l)\) denote the projected residual-stream
contribution of attention head \(h\) in layer \(l\), where \(\mathbf{S}_l\) is the sequence of hidden states at layer \(l\), and let
\(n_{l,j}(\mathbf{s}_l)\) denote the contribution of MLP intermediate
channel \(j\), including its outgoing down projection, where \(\mathbf{s}_l\) is the per-token hidden state.
The masked module outputs are:
\begin{equation}
\left\{
\begin{aligned}
\mathrm{MHA}^{M}_l(\mathbf{S}_l)
  &= \sum_h m^a_{l,h}\,a_{l,h}(\mathbf{S}_l),\\
\mathrm{MLP}^{M}_l(\mathbf{s}_l)
  &= \sum_j m^n_{l,j}\,n_{l,j}(\mathbf{s}_l),
\end{aligned}
\right.
\label{eq:component_masks}
\end{equation}
where \(m^a_{l,h}=m(\alpha_{l,h})\) and \(m^n_{l,j}=m(\beta_{l,j})\)
are obtained from learnable logits through a straight-through
binarization, with \(\sigma\) denoting the sigmoid function:
\begin{equation}
m(t)=\begin{cases}
\mathbb{1}[\sigma(t)>0.5] & \text{forward pass,}\\
\sigma(t) & \text{gradient computation.}
\end{cases}
\label{eq:binarization}
\end{equation}
This produces binary gates during inference while allowing
gradient-based optimization of the logits.
The retained branch \(\mathcal{G}_{M}\) uses gates \(M\) and learns
reference guard behavior, while the complementary branch
\(\mathcal{G}_{1-M}\) replaces each gate with \(1-m\) and learns the
observed shortcut behavior.
All remaining model components are shared between the two branches.

\paragraph{Intervention examples and targets.}
Let
\(x_{\mathrm{train}}=(u_{\mathrm{train}},r_{\mathrm{train}})\)
denote an example from the training dataset.
For \(p\in\mathcal{P}_3\), a ground-truth harmful sample--cue pair
\((x_{\mathrm{train}},p)\) is considered shortcut-sensitive when:
\begin{equation}
\left\{
\begin{aligned}
g_H(x_{\mathrm{train}})&=H,\\
g_H(T_{p,\mathrm{head}}(x_{\mathrm{train}}))&=U.
\end{aligned}
\right.
\label{eq:shortcut_sensitive}
\end{equation}
All cues used for intervention-example construction are inserted at the
response head.
We select 300 shortcut-sensitive sample-cue pairs from WildGuardMix
and generate all optimization targets using the original guard.

For direct-output guards, each cue-inserted input and its original
guard output define a shortcut-target pair
\((x^{\mathrm{sc}},y_{\mathrm{sc}})\) for the complementary branch.
The correct-target term uses 75 original refusal examples and 75
original compliance examples, balanced to preserve both behaviors.
Each input is paired with its original guard output to form
\((x^{\mathrm{corr}},y_{\mathrm{corr}})\) for the retained branch.

For GuardReasoner, which generates a reasoning trace before the final prediction, directly contrasting the complete no-cue and cue-conditioned outputs would introduce differences in the reasoning traces.
We therefore construct a cross-prefix contrast from two cue insertions
applied to the same underlying response: one preserves the correct
harmful prediction, whereas the other induces a shortcut-driven flip to
unharmful.
The corresponding model-generated reasoning traces and target decisions
define \((x^{\mathrm{corr}},y_{\mathrm{corr}})\) and
\((x^{\mathrm{sc}},y_{\mathrm{sc}})\), respectively.
Because both inputs contain a refusal cue, the contrast reduces the influence of cue presence and focuses the optimization on the shortcut-induced prediction flip.
Balanced refusal and compliance examples are also included in the
correct-target term to discourage disruption of legitimate refusal
detection.
All output tokens contribute to the loss.

After optimization, components assigned to the complementary branch are
permanently suppressed by zeroing the corresponding projection slices.
The resulting guard retains only the selected branch and requires neither
a mask wrapper nor an additional inference pass.

\paragraph{Optimization objective.}
Given the correct-target and shortcut-target training pairs
\((x_{\mathrm{train}}^{\mathrm{corr}},y_{\mathrm{corr}})\) and
\((x_{\mathrm{train}}^{\mathrm{sc}},y_{\mathrm{sc}})\), respectively,
we optimize:
\begin{equation}
\mathcal{L}
=
\lambda_{\mathrm{corr}}\mathcal{L}_{\mathrm{corr}}(M)
+
\lambda_{\mathrm{sc}}\mathcal{L}_{\mathrm{sc}}(1-M)
+
\gamma\mathcal{L}_{\mathrm{sparse}},
\label{eq:mask_objective}
\end{equation}
where
\(\mathcal{L}_{\mathrm{corr}}(M)
=\mathrm{CE}(\mathcal{G}_{M}(x_{\mathrm{train}}^{\mathrm{corr}}),
y_{\mathrm{corr}})\)
and
\(\mathcal{L}_{\mathrm{sc}}(1-M)
=\mathrm{CE}(\mathcal{G}_{1-M}(x_{\mathrm{train}}^{\mathrm{sc}}),
y_{\mathrm{sc}})\).
The sparsity term penalizes the average gate mass assigned to the
complementary branch, normalized separately over attention heads and MLP
neurons so that neither component type dominates because of its size.
The weights \(\lambda_{\mathrm{corr}}\) and
\(\lambda_{\mathrm{sc}}\) balance the two target objectives, while
\(\gamma\) controls the sparsity of the complementary branch.
All original model parameters remain frozen.

\paragraph{Mitigation setup.}
We evaluate mitigation on seven guards, excluding the two Aegis2-derived guards because of their limited baseline vulnerability.
We retain the comparatively robust Qwen3Guard-8B to enable a within-family comparison across model sizes.
By default, each mask is optimized on selected examples from WildGuardMix for 50 epochs, with
\(\lambda_{\mathrm{corr}}=10\),
\(\lambda_{\mathrm{sc}}=1\), and
\(\gamma=10\).
Mitigation is evaluated primarily on WildGuardTest and Aegis2 Test.
All original guard parameters remain frozen, and mask optimization requires only a single A100 80GB GPU.
Implementation details and additional cross-dataset results on BeaverTails~\cite{ji2023beavertails} are provided in Appendix.

\subsection{Q1: Generalizability of Sparse Component Suppression}

\paragraph{Suppression at the optimized position.}
Table~\ref{tab:mitigation} compares original and masked guards.
On WildGuardTest, mean head-position \(\dfr@3\) drops from 16.39\% to 3.44\% (79\% relative reduction); on Aegis2 Test, from 20.35\% to 4.14\% (80\% relative reduction).
The reduction is consistent across guards: all seven show lower head-position \(\dfr@3\) after masking on both evaluation sets.
WG-7B exhibits the largest absolute reduction, with head-position \(\dfr@3\) falling from 37.72\% to 2.85\% on WildGuardTest and from 54.82\% to 4.82\% on Aegis2 Test.
The two GR variants retain the highest post-masking head-position \(\dfr@3\) on both datasets.

\paragraph{Transfer across positions and datasets.}
Component masks are optimized on head-position examples only, yet the effect generalizes to unseen positions.
On WildGuardTest, mean middle- and tail-position \(\dfr@3\) decrease from 7.66\% and 12.17\% to 2.28\% and 2.82\%, respectively.
On Aegis2 Test, the corresponding values decrease from 19.73\% and 25.34\% to 5.23\% and 6.31\%.
Across all position comparisons, the sole increase occurs for GR-1B on the WildGuardTest tail, where \(\dfr@3\) rises marginally from 5.34\% to 5.67\%.
On additional BeaverTails dataset, mean head-position \(\dfr@3\) also decreases from 17.00\% to 4.00\% (shown in Appendix).
As suppressing the same guard-specific components remains effective across cue positions and held-out datasets, refusal-cue failures across these settings appear to be partly mediated by a shared set of internal components.

\paragraph{Clean harmfulness classification is broadly retained.}
On WildGuardTest, mean harmfulness F1 decreases by only 0.30\%, and harmful recall improves for six of seven models. QG-0.6B is the sole exception (\(-3.20\%\)).
On Aegis2 Test, mean harmfulness F1 increases by 2.51\%.
Sparse component suppression thus reduces the refusal-cue shortcut across guards, cue positions, and datasets while broadly preserving clean harmfulness classification.

\begin{table}[htbp]
\centering
\small
\setlength{\tabcolsep}{3pt}
\begin{tabular}{llrrrr}
\toprule
Model & State
& Ref. F1\(\uparrow\)
& Comp. R\(\uparrow\)
& Ref. R\(\uparrow\)
& \(\rfr@3\downarrow\) \\
\midrule
\multirow{2}{*}{WG-7B}
 & Orig. & 88.54 & 88.62 & 98.01 & 51.52 \\
 & Mask  & 90.18 & 93.47 & 93.12 & \textbf{2.93} \\
\midrule
\multirow{2}{*}{GR-1B}
 & Orig. & 89.09 & 88.72 & 98.92 & 30.35 \\
 & Mask  & 87.61 & 87.57 & 97.83 & \textbf{15.50} \\
\midrule
\multirow{2}{*}{GR-8B}
 & Orig. & 90.07 & 89.69 & 99.28 & 13.96 \\
 & Mask  & 90.53 & 90.66 & 98.55 & \textbf{8.92} \\
\midrule
\multirow{2}{*}{QG-0.6B}
 & Orig. & 82.95 & 80.97 & 98.55 & 17.06 \\
 & Mask  & 82.04 & 81.59 & 95.84 & \textbf{2.72} \\
\midrule
\multirow{2}{*}{QG-8B}
 & Orig. & 84.64 & 82.56 & 99.64 & 13.45 \\
 & Mask  & 82.53 & 80.00 & 99.10 & \textbf{8.22} \\
\midrule
\multirow{2}{*}{\textbf{Macro avg.}}
 & Orig. & 87.06 & 86.11 & 98.88 & 25.27 \\
 & Mask  & 86.58 & 86.66 & 96.89 & \textbf{7.66} \\
\bottomrule
\end{tabular}
\caption{
Clean refusal recognition and cue-induced compliance-to-refusal flips
before and after component suppression.
Ref. F1, Comp. R, and Ref. R denote refusal F1, compliance recall,
and refusal recall on unmodified responses, respectively.
}
\label{tab:refusal_det}
\end{table}



\subsection{Q2: Partial Functional Separability from Refusal Recognition}

\paragraph{Clean refusal recognition is broadly preserved.}
Table~\ref{tab:refusal_det} reports clean refusal metrics and cue-induced refusal flips for the five guards with an explicit refusal output.
After component suppression, refusal F1 changes by at most 2.11\%.
Mean compliance recall increases slightly from 86.11\% to 86.66\%, while mean refusal recall decreases by 1.99 percentage points from 98.88\% to 96.89\%, reflecting a mild class-wise shift rather than uniform degradation.

\paragraph{The two behaviors are partially separable under component-level intervention.}
Component suppression reduces mean \(\rfr@3\) from 25.27\% to 7.66\%, with reductions for every guard, yet clean refusal F1 and recall change only marginally.
This dissociation, together with the partial flip coupling observed before masking (Section~\ref{sec:flip_coupling}), provides intervention-based evidence that shortcut reliance can be suppressed at the component level without disrupting legitimate refusal recognition.


Ablation studies on component granularity and sparsity weight are reported in Appendix.

\section{Conclusion}

This work identifies the refusal-cue shortcut in response-level safety guards and shows that it is widespread but heterogeneous across model families and response positions, with smaller variants generally exhibiting greater vulnerability.
Sparse component suppression substantially reduces cue-induced failures while broadly preserving clean harmfulness performance and legitimate refusal recognition, providing intervention-based evidence that shortcut reliance and refusal recognition are partially separable.
By revealing a systematic weakness in current safety guards and offering a lightweight post-training mitigation, this work contributes to the development of more reliable content-safety systems.
Our evaluation primarily relies on widely used safety benchmarks, which provide controlled and systematic settings but may not capture the diversity and complexity of real-world applications. Future work should examine shortcut prevalence and mitigation effectiveness in broader domain-specific and deployment scenarios~\cite{rottger2024xstest,mazeika2024harmbench}.

{
    \small
    \bibliographystyle{ieeenat_fullname}
    \bibliography{main}

@article{yang2025qwen3,
  title={Qwen3 technical report},
  author={Yang, An and Li, Anfeng and Yang, Baosong and Zhang, Beichen and Hui, Binyuan and Zheng, Bo and Yu, Bowen and Gao, Chang and Huang, Chengen and Lv, Chenxu and others},
  journal={arXiv preprint arXiv:2505.09388},
  year={2025}
}

@article{chen2021evaluating,
  title={Evaluating large language models trained on code},
  author={Chen, Mark and Tworek, Jerry and Jun, Heewoo and Yuan, Qiming and Pinto, Henrique Ponde De Oliveira and Kaplan, Jared and Edwards, Harri and Burda, Yuri and Joseph, Nicholas and Brockman, Greg and others},
  journal={arXiv preprint arXiv:2107.03374},
  year={2021}
}

@article{zhang2025stair,
  title={Stair: Improving safety alignment with introspective reasoning},
  author={Zhang, Yichi and Zhang, Siyuan and Huang, Yao and Xia, Zeyu and Fang, Zhengwei and Yang, Xiao and Duan, Ranjie and Yan, Dong and Dong, Yinpeng and Zhu, Jun},
  journal={arXiv preprint arXiv:2502.02384},
  year={2025}
}

@article{zhang2025alphaalign,
  title={AlphaAlign: Incentivizing Safety Alignment with Extremely Simplified Reinforcement Learning},
  author={Zhang, Yi and Zhang, An and Zhang, XiuYu and Sheng, Leheng and Chen, Yuxin and Liang, Zhenkai and Wang, Xiang},
  journal={arXiv preprint arXiv:2507.14987},
  year={2025}
}

@article{ji2023beavertails,
  title={Beavertails: Towards improved safety alignment of llm via a human-preference dataset},
  author={Ji, Jiaming and Liu, Mickel and Dai, Josef and Pan, Xuehai and Zhang, Chi and Bian, Ce and Chen, Boyuan and Sun, Ruiyang and Wang, Yizhou and Yang, Yaodong},
  journal={Advances in Neural Information Processing Systems},
  volume={36},
  pages={24678--24704},
  year={2023}
}

@article{han2024wildguard,
  title={Wildguard: Open one-stop moderation tools for safety risks, jailbreaks, and refusals of llms},
  author={Han, Seungju and Rao, Kavel and Ettinger, Allyson and Jiang, Liwei and Lin, Bill Yuchen and Lambert, Nathan and Choi, Yejin and Dziri, Nouha},
  journal={Advances in Neural Information Processing Systems},
  volume={37},
  pages={8093--8131},
  year={2024}
}

@article{liu2025guardreasoner,
  title={Guardreasoner: Towards reasoning-based llm safeguards},
  author={Liu, Yue and Gao, Hongcheng and Zhai, Shengfang and He, Yufei and Xia, Jun and Hu, Zhengyu and Chen, Yulin and Yang, Xihong and Zhang, Jiaheng and Li, Stan Z and others},
  journal={arXiv preprint arXiv:2501.18492},
  year={2025}
}

@inproceedings{ghosh2025aegis2,
  title={Aegis2. 0: A diverse ai safety dataset and risks taxonomy for alignment of llm guardrails},
  author={Ghosh, Shaona and Varshney, Prasoon and Sreedhar, Makesh Narsimhan and Padmakumar, Aishwarya and Rebedea, Traian and Varghese, Jibin Rajan and Parisien, Christopher},
  booktitle={Proceedings of the 2025 Conference of the Nations of the Americas Chapter of the Association for Computational Linguistics: Human Language Technologies (Volume 1: Long Papers)},
  pages={5992--6026},
  year={2025}
}

@article{sreedhar2025safety,
  title={Safety through reasoning: An empirical study of reasoning guardrail models},
  author={Sreedhar, Makesh Narsimhan and Rebedea, Traian and Parisien, Christopher},
  journal={Findings of the Association for Computational Linguistics: EMNLP},
  volume={2025},
  pages={21862--21880},
  year={2025}
}

@article{grattafiori2024llama,
  title={The llama 3 herd of models},
  author={Grattafiori, Aaron and Dubey, Abhimanyu and Jauhri, Abhinav and Pandey, Abhinav and Kadian, Abhishek and Al-Dahle, Ahmad and Letman, Aiesha and Mathur, Akhil and Schelten, Alan and Vaughan, Alex and others},
  journal={arXiv preprint arXiv:2407.21783},
  year={2024}
}

@article{inan2023llama,
  title={Llama guard: Llm-based input-output safeguard for human-ai conversations},
  author={Inan, Hakan and Upasani, Kartikeya and Chi, Jianfeng and Rungta, Rashi and Iyer, Krithika and Mao, Yuning and Tontchev, Michael and Hu, Qing and Fuller, Brian and Testuggine, Davide and others},
  journal={arXiv preprint arXiv:2312.06674},
  year={2023}
}

@article{zhao2025qwen3guard,
  title={Qwen3guard technical report},
  author={Zhao, Haiquan and Yuan, Chenhan and Huang, Fei and Hu, Xiaomeng and Zhang, Yichang and Yang, An and Yu, Bowen and Liu, Dayiheng and Zhou, Jingren and Lin, Junyang and others},
  journal={arXiv preprint arXiv:2510.14276},
  year={2025}
}

@article{geirhos2020shortcut,
  title={Shortcut learning in deep neural networks},
  author={Geirhos, Robert and Jacobsen, J{\"o}rn-Henrik and Michaelis, Claudio and Zemel, Richard and Brendel, Wieland and Bethge, Matthias and Wichmann, Felix A},
  journal={Nature Machine Intelligence},
  volume={2},
  number={11},
  pages={665--673},
  year={2020},
  publisher={Nature Publishing Group UK London}
}

@inproceedings{conmy2023automated,
  title={Towards Automated Circuit Discovery for Mechanistic Interpretability},
  author={Conmy, Arthur and Mavor-Parker, Augustine N. and Lynch, Aengus and Heimersheim, Stefan and Garriga-Alonso, Adri{\`a}},
  booktitle={Advances in Neural Information Processing Systems},
  volume={36},
  year={2023},
}

@article{li2023inference,
  title={Inference-time intervention: Eliciting truthful answers from a language model},
  author={Li, Kenneth and Patel, Oam and Vi{\'e}gas, Fernanda and Pfister, Hanspeter and Wattenberg, Martin},
  journal={Advances in Neural Information Processing Systems},
  volume={36},
  pages={41451--41530},
  year={2023}
}

@article{yu2026safeseek,
  title={SafeSeek: Universal Attribution of Safety Circuits in Language Models},
  author={Yu, Miao and Fu, Siyuan and Aloqaily, Moayad and Zhou, Zhenhong and Otoum, Safa and Wang, Kun and Guo, Yufei and Wen, Qingsong and others},
  journal={arXiv preprint arXiv:2603.23268},
  year={2026}
}

@inproceedings{tasawong2025shortcut,
  title={Shortcut Learning in Safety: The Impact of Keyword Bias in Safeguards},
  author={Tasawong, Panuthep and Laosaengpha, Napat and Ponwitayarat, Wuttikorn and Lim, Sitiporn and Manakul, Potsawee and Cahyawijaya, Samuel and Udomcharoenchaikit, Can and Limkonchotiwat, Peerat and Chuangsuwanich, Ekapol and Nutanong, Sarana},
  booktitle={Proceedings of the The First Workshop on LLM Security (LLMSEC)},
  pages={189--197},
  year={2025}
}

@inproceedings{qi2025safety,
  title={Safety alignment should be made more than just a few tokens deep},
  author={Qi, Xiangyu and Panda, Ashwinee and Lyu, Kaifeng and Ma, Xiao and Roy, Subhrajit and Beirami, Ahmad and Mittal, Prateek and Henderson, Peter},
  booktitle={International Conference on Learning Representations},
  volume={2025},
  pages={54911--54941},
  year={2025}
}

@inproceedings{yuan2025refuse,
  title={Refuse whenever you feel unsafe: Improving safety in llms via decoupled refusal training},
  author={Yuan, Youliang and Jiao, Wenxiang and Wang, Wenxuan and Huang, Jen-tse and Xu, Jiahao and Liang, Tian and He, Pinjia and Tu, Zhaopeng},
  booktitle={Proceedings of the 63rd Annual Meeting of the Association for Computational Linguistics (Volume 1: Long Papers)},
  pages={3149--3167},
  year={2025}
}

@article{zhao2026llms,
  title={Llms encode harmfulness and refusal separately},
  author={Zhao, Jiachen and Huang, Jing and Wu, Zhengxuan and Bau, David and Shi, Weiyan},
  journal={Advances in Neural Information Processing Systems},
  volume={38},
  pages={140283--140318},
  year={2026}
}

@misc{team2024gemma,
  title={Gemma 2: Improving Open Language Models at a Practical Size},
  author={{Gemma Team} and Morgane Riviere and Shreya Pathak and Pier Giuseppe Sessa and others},
  year={2024},
  eprint={2408.00118},
  archivePrefix={arXiv},
  primaryClass={cs.CL},
}

@article{sagawa2019distributionally,
  title={Distributionally robust neural networks for group shifts: On the importance of regularization for worst-case generalization},
  author={Sagawa, Shiori and Koh, Pang Wei and Hashimoto, Tatsunori B and Liang, Percy},
  journal={arXiv preprint arXiv:1911.08731},
  year={2019}
}

@article{ouyang2022training,
  title={Training language models to follow instructions with human feedback},
  author={Ouyang, Long and Wu, Jeff and Jiang, Xu and Almeida, Diogo and Wainwright, Carroll L and Mishkin, Pamela and Zhang, Chong and Agarwal, Sandhini and Slama, Katarina and Ray, Alex and others},
  journal={arXiv preprint arXiv:2203.02155},
  year={2022}
}

@article{bai2022training,
  title={Training a helpful and harmless assistant with reinforcement learning from human feedback},
  author={Bai, Yuntao and Jones, Andy and Ndousse, Kamal and Askell, Amanda and Chen, Anna and DasSarma, Nova and Drain, Dawn and Fort, Stanislav and Ganguli, Deep and Henighan, Tom and others},
  journal={arXiv preprint arXiv:2204.05862},
  year={2022}
}

@article{arditi2024refusal,
  title={Refusal in language models is mediated by a single direction},
  author={Arditi, Andy and Obeso, Oscar and Syed, Aaquib and Paleka, Daniel and Panickssery, Nina and Gurnee, Wes and Nanda, Neel},
  journal={Advances in Neural Information Processing Systems},
  volume={37},
  pages={136037--136083},
  year={2024}
}

@article{wei2023jailbroken,
  title={Jailbroken: How does llm safety training fail?},
  author={Wei, Alexander and Haghtalab, Nika and Steinhardt, Jacob},
  journal={Advances in Neural Information Processing Systems},
  volume={36},
  pages={80079--80110},
  year={2023}
}

@inproceedings{rottger2024xstest,
  title={Xstest: A test suite for identifying exaggerated safety behaviours in large language models},
  author={R{\"o}ttger, Paul and Kirk, Hannah and Vidgen, Bertie and Attanasio, Giuseppe and Bianchi, Federico and Hovy, Dirk},
  booktitle={Proceedings of the 2024 Conference of the North American Chapter of the Association for Computational Linguistics: Human Language Technologies (Volume 1: Long Papers)},
  pages={5377--5400},
  year={2024}
}

@article{yuan2024seval,
  title={S-eval: Towards automated and comprehensive safety evaluation for large language models},
  author={Yuan, Xiaohan and Li, Jinfeng and Wang, Dongxia and Chen, Yuefeng and Mao, Xiaofeng and Huang, Longtao and Chen, Jialuo and Xue, Hui and Liu, Xiaoxia and Wang, Wenhai and others},
  journal={arXiv preprint arXiv:2405.14191},
  year={2024}
}

@inproceedings{krasnodkebska2026safety,
  title={Safety of large language models beyond English: A systematic literature review of risks, biases, and safeguards},
  author={Krasnodębska, Aleksandra and Dziewulska, Katarzyna and Seweryn, Karolina and Chrabaszcz, Maciej and Kusa, Wojciech},
  booktitle={Proceedings of the 19th Conference of the European Chapter of the Association for Computational Linguistics (Volume 1: Long Papers)},
  pages={1003--1034},
  year={2026}
}

@article{mazeika2024harmbench,
  title={Harmbench: A standardized evaluation framework for automated red teaming and robust refusal},
  author={Mazeika, Mantas and Phan, Long and Yin, Xuwang and Zou, Andy and Wang, Zifan and Mu, Norman and Sakhaee, Elham and Li, Nathaniel and Basart, Steven and Li, Bo and others},
  journal={arXiv preprint arXiv:2402.04249},
  year={2024}
}
}

\newpage
\section*{Appendix}           
\appendix

This appendix reports dataset descriptions, training dataset audit counts, individual cue and control results, query-budget scaling, BeaverTails transfer, and detailed intervention optimization settings.
All reported metrics are percentages.
Detection Failure Rate (\(\dfr\)) and Refusal Flip Rate (\(\rfr\)) use the complete ground-truth harmful subset as their fixed denominator, identical to harmful recall, as defined in the main paper.
\(\dfr@1\) reports one cue, while \(\dfr@3\) is the sample-level union over R1, R2, and R\textsubscript{greet}.
The corresponding control union over N1, S1, and Rand is denoted \(\cdf@3\).

\section{Evaluation Datasets}
\label{app:datasets}

We use three evaluation datasets covering distinct construction methodologies and safety taxonomies.

\paragraph{WildGuardTest.}
WildGuardTest is the test split of the WildGuardMix benchmark~\cite{han2024wildguard}, originally containing 1,725 prompt-response pairs spanning 13 harm categories.
Each example is annotated with three labels: prompt harmfulness, response harmfulness, and response refusal.
The response-level refusal labels enable joint evaluation of harmfulness classification and refusal recognition.
We require all three annotation fields to be present and exclude 37 samples with at least one missing label (16 missing response harmfulness, 26 missing prompt harmfulness, and 5 missing response refusal, with partial overlap), yielding 1,688 evaluation examples comprising 282 harmful and 1,406 unharmful responses.
The harmful subset (282 examples) serves as the fixed denominator for DFR and RFR, while harmfulness F1 is computed over all 1,688 examples.
WildGuardTest serves as the primary evaluation set and shares its taxonomy and annotation schema with the WildGuardMix training data used for mask optimization, making it an in-distribution evaluation.

\paragraph{Aegis2 Test.}
Aegis2 Test is the evaluation split of the Aegis2.0 safety dataset~\cite{ghosh2025aegis2}, which covers a 14-category risk taxonomy including critical safety, content safety, and societal risks.
Unlike WildGuardMix, Aegis2 augments its training data with synthetic safe responses that go beyond simple refusals to include alternative assistance, harm explanation, and conversational redirection.
The original test split contains 1,964 examples, of which 1,112 carry only a prompt-level label and lack a response-level annotation.
We retain the 852 examples that have both prompt and response labels, comprising 394 harmful and 458 unharmful responses.
DFR is computed over the 394 harmful examples, while harmfulness F1 is computed over all 852 examples.

\paragraph{BeaverTails.}
BeaverTails~\cite{ji2023beavertails} is a large-scale safety meta-dataset originally designed for preference-based safety alignment.
It provides binary safety labels across 14 harm categories for over 330K QA pairs.
We evaluate on a test split containing 3,021 examples comprising 1,733 harmful and 1,288 unharmful responses.
DFR is computed over the 1,733 harmful examples, while harmfulness F1 is computed over all 3,021 examples.
BeaverTails is not used during mask optimization and differs from the two primary datasets in both annotation schema and content distribution, making it a fully held-out cross-dataset evaluation.

\section{Training Data Audit Details}
\label{app:data_audit}

Table~\ref{tab:data_dist} reports the response-pattern counts used in Figure~1(a) of the main paper.
The audit is restricted to responses associated with harmful prompts.
Percentages are computed within each displayed dataset subset.
WildGuardMix contains no refusal+harmful example.
GR-Train contains 114 such examples, which account for 0.26\% of the displayed subset and 0.58\% of all refusal responses in that subset.
These counts are obtained after cross-checking each sample's chain-of-thought reasoning against its final label and retaining only those where the two are consistent

\begin{table}[t]
\centering
\small
\setlength{\tabcolsep}{2.2pt}
\begin{tabular}{lrrrr}
\toprule
& \multicolumn{2}{c}{WildGuardMix} & \multicolumn{2}{c}{GR-Train} \\
\cmidrule(lr){2-3} \cmidrule(lr){4-5}
Response pattern & Count & \% & Count & \% \\
\midrule
Refusal + Unharmful & 10{,}651 & 50.04 & 19{,}636 & 45.59 \\
Compliance + Unharmful & 2{,}294 & 10.78 & 2{,}833 & 6.58 \\
Compliance + Harmful & 8{,}341 & 39.19 & 20{,}491 & 47.57 \\
\textbf{Refusal + Harmful} & \textbf{0} & \textbf{0.00} & \textbf{114} & \textbf{0.26} \\
\midrule
Total & 21{,}286 & 100 & 43{,}074 & 100 \\
\bottomrule
\end{tabular}
\caption{Distribution of response labels under harmful prompts.}
\label{tab:data_dist}
\end{table}

\section{Cue and Control Results}
\label{app:cue_controls}

\subsection{Individual Refusal Cues}

Table~\ref{tab:individual_cues} expands the three primary cues with three additional refusal formulations.
The results are reported as \(\dfr@1\), so each column measures the dataset-level rate of flips from harmful to unharmful induced by one head-position cue.
The additional formulations produce nonzero failures across all seven reported models.
Their relative strength varies by architecture, which is consistent with reliance on a broader refusal feature rather than one universal string. The refusal cues are shown as follows:
\begin{itemize}
\setlength{\itemsep}{0pt}
\setlength{\parsep}{0pt}
\setlength{\topsep}{2pt}
  \item \textbf{R1}: \textit{I cannot fulfill your request.}
  \item \textbf{R2}: \textit{I apologize, but I cannot fulfill your request.}
  \item \textbf{R\textsubscript{greet}}: \textit{Hello! I'm here to help, but I cannot assist with that request.}
  \item \textbf{R3}: \textit{I'm sorry, but I cannot fulfill your request.}
  \item \textbf{R\textsubscript{role}}: \textit{As a helpful and respectful assistant, I cannot help with this.}
  \item \textbf{R\textsubscript{provide}}: \textit{I cannot provide information or guidance on this topic.}
\end{itemize}

\begin{table}[t]
\centering
\small
\setlength{\tabcolsep}{3.0pt}
\begin{tabular}{lcccccc}
\toprule
Model & R1\(\downarrow\) & R2\(\downarrow\) & R\textsubscript{greet}\(\downarrow\) & R3\(\downarrow\) & R\textsubscript{role}\(\downarrow\) & R\textsubscript{provide}\(\downarrow\) \\
\midrule
WG-7B     & 21.35 & 37.72 & 20.28 & 37.72 & 14.95 & 7.12 \\
GR-1B     &  9.96 & 13.17 & 17.44 & 16.01 & 17.79 & 18.86 \\
GR-8B     &  9.22 & 11.70 &  8.87 & 11.35 &  9.22 & 9.22 \\
LG3-1B    & 16.67 & 15.96 & 15.96 & 16.67 & 19.50 & 15.96 \\
LG3-8B    &  6.03 &  5.32 &  3.55 &  5.32 &  2.84 & 3.55 \\
QG-0.6B   &  4.26 &  7.80 &  8.87 &  9.22 & 13.83 & 6.03 \\
QG-8B     &  0.35 &  4.26 &  4.61 &  4.26 &  8.16 & 1.42 \\
\bottomrule
\end{tabular}
\caption{Head-position \(\dfr@1\) on WildGuardTest for six refusal formulations: three primary cues (R1, R2, R\textsubscript{greet}) and three additional variants (R3, R\textsubscript{role}, R\textsubscript{provide}). All values are percentages.}
\label{tab:individual_cues}
\end{table}

\subsection{Individual Controls}

The three controls use exactly the same head-position insertion logic as the refusal cues.
Table~\ref{tab:individual_controls} reports both their individual \(\dfr@1\) values and the sample-level control union.
For WG-7B and both GR models, \(\cdf@3\) remains substantially below the corresponding refusal-cue \(\dfr@3\) in the main paper.
For the Aegis2-derived models, the two unions are closer, consistent with their lower refusal-specific vulnerability.

\begin{table}[t]
\centering
\small
\setlength{\tabcolsep}{3.0pt}
\begin{tabular}{lrrrr}
\toprule
Model & N1\(\downarrow\) & S1\(\downarrow\) & Rand\(\downarrow\) & \(\cdf@3\downarrow\) \\
\midrule
WG-7B      & 0.00 & 0.71 & 0.71 & 0.71 \\
GR-1B      & 0.36 & 2.85 & 3.20 & 4.98 \\
GR-8B      & 2.48 & 4.26 & 2.13 & 5.67 \\
LNSGV2-8B  & 5.06 & 2.13 & 4.96 & 5.28 \\
NCSR-4B    & 1.77 & 1.42 & 1.42 & 3.55 \\
LG3-1B     & 0.71 & 2.48 & 2.48 & 4.61 \\
LG3-8B     & 1.42 & 2.13 & 1.42 & 4.26 \\
QG-0.6B    & 0.71 & 4.61 & 1.77 & 4.96 \\
QG-8B      & 0.00 & 3.90 & 0.71 & 3.90 \\
\bottomrule
\end{tabular}
\caption{Individual control \(\dfr@1\) and the three-control union on WildGuardTest. All values are percentages.}
\label{tab:individual_controls}
\end{table}

\section{Scaling the Cue Query Budget}
\label{app:k_scaling}

\(\dfr@3\) represents a fixed query budget rather than an upper bound over all possible refusal cues.
Table~\ref{tab:k_scaling} accumulates cues in the fixed order
\[
\begin{aligned}
\mathrm{R1}&\rightarrow\mathrm{R2}\rightarrow\mathrm{R}_{\mathrm{greet}}\\
&\rightarrow\mathrm{R3}\rightarrow\mathrm{R}_{\mathrm{role}}
\rightarrow\mathrm{R}_{\mathrm{provide}}.
\end{aligned}
\]
The \(K=3\) column exactly matches the primary head-position \(\dfr@3\).
WG-7B saturates after the second cue, and both GR models continue to gain failures as the budget grows.

\begin{table}[t]
\centering
\small
\setlength{\tabcolsep}{2.0pt}
\begin{tabular}{lrrrrrr}
\toprule
& \multicolumn{6}{c}{\(\dfr@K\downarrow\)} \\
\cmidrule(lr){2-7}
Model & \(K=1\) & \(K=2\) & \(K=3\) & \(K=4\) & \(K=5\) & \(K=6\) \\
\midrule
WG-7B & 21.35 & 37.72 & \textbf{37.72} & 38.08 & 38.08 & 38.08 \\
GR-1B &  9.96 & 15.30 & \textbf{21.71} & 23.84 & 25.98 & 28.11 \\
GR-8B &  9.22 & 14.18 & \textbf{15.96} & 16.67 & 17.02 & 18.44 \\
\bottomrule
\end{tabular}
\caption{Sample-level \(\dfr@K\) under a fixed cue accumulation order. All values are percentages.}
\label{tab:k_scaling}
\end{table}

\section{BeaverTails Transfer}
\label{app:beavertails}

BeaverTails~\cite{ji2023beavertails} contains 3,021 test examples (1,733 harmful and 1,288 unharmful responses) and is not used to optimize the component masks.
Table~\ref{tab:beaver} evaluates the same three primary refusal cues used in the main paper.
Mean head, middle, and tail \(\dfr@3\) decrease from 17.00\%, 16.76\%, and 18.30\% to 4.00\%, 4.20\%, and 4.33\%, respectively.
Mean harmfulness F1 increases from 79.88\% to 82.25\%.
The largest reduction occurs for WG-7B.
Residual failures remain higher for GR-1B and GR-8B, matching the model-level pattern on the two primary datasets.
The result supports cross-dataset transfer of the intervention without implying complete removal of the shortcut.

\begin{table*}[t]
\centering
\small
\setlength{\tabcolsep}{4.0pt}
\begin{tabular}{llcccccc}
\toprule
Model & State & Harm. F1\(\uparrow\) & Harm. R\(\uparrow\) & Head\(\downarrow\) & Middle\(\downarrow\) & Tail\(\downarrow\) & Mask\% \\
\midrule
\multirow{2}{*}{WG-7B}
 & Orig. & 84.11 & 79.28 & 45.24 & 40.57 & 36.24 & -- \\
 & Mask  & 84.67 & 82.86 &  5.14 &  4.39 &  3.52 & 8.6 \\
\midrule
\multirow{2}{*}{GR-1B}
 & Orig. & 85.29 & 82.80 & 17.60 & 13.91 & 11.60 & -- \\
 & Mask  & 84.25 & 79.92 & 11.14 & 11.25 &  9.46 & 2.7 \\
\midrule
\multirow{2}{*}{GR-8B}
 & Orig. & 87.86 & 87.36 & 11.31 &  6.92 &  5.60 & -- \\
 & Mask  & 87.60 & 88.23 &  7.10 &  5.08 &  4.85 & 4.3 \\
\midrule
\multirow{2}{*}{LG3-1B}
 & Orig. & 63.62 & 47.78 & 17.02 & 24.81 & 39.41 & -- \\
 & Mask  & 71.98 & 60.24 &  1.67 &  3.46 &  4.96 & 1.6 \\
\midrule
\multirow{2}{*}{LG3-8B}
 & Orig. & 67.81 & 52.63 & 10.68 &  9.75 & 13.27 & -- \\
 & Mask  & 75.36 & 64.05 &  0.23 &  0.92 &  2.37 & 1.9 \\
\midrule
\multirow{2}{*}{QG-0.6B}
 & Orig. & 84.67 & 79.23 & 12.00 & 16.50 & 17.02 & -- \\
 & Mask  & 85.58 & 83.04 &  0.98 &  2.02 &  2.42 & 0.9 \\
\midrule
\multirow{2}{*}{QG-8B}
 & Orig. & 85.80 & 82.29 &  5.14 &  4.85 &  4.96 & -- \\
 & Mask  & 86.34 & 83.67 &  1.73 &  2.25 &  2.71 & 0.7 \\
\midrule
\multirow{2}{*}{\textbf{Macro avg.}}
 & Orig. & 79.88 & 73.05 & 17.00 & 16.76 & 18.30 & -- \\
 & Mask  & 82.25 & 77.43 & \textbf{4.00} & \textbf{4.20} & \textbf{4.33} & -- \\
\bottomrule
\end{tabular}
\caption{Original and masked guard performance on BeaverTails. Harm. F1 and Harm. R denote response harmfulness F1 and harmful recall on unmodified responses. Head, Middle, and Tail report \(\dfr@3\) under short refusal cues. All values are percentages.}
\label{tab:beaver}
\end{table*}

\section{Intervention Optimization Details}
\label{app:hyperparams}

\paragraph{Training data construction.}
Intervention examples are constructed from WildGuardMix training data.
For each guard, we first run the three primary refusal cues (R1, R2, R\textsubscript{greet}) at the response head on all ground-truth harmful examples and identify shortcut-sensitive pairs, i.e., those for which the original guard predicts harmful but the cue-inserted version predicts unharmful.
We sample \(N=300\) shortcut-sensitive pairs after deduplication by sample ID: one random prefix per sample is retained first, and additional prefix entries from already-seen samples are used as backfill if fewer than 300 unique samples are available.
To preserve legitimate refusal detection, 150 additional examples (75 refusal + 75 compliance, stratified) from the WildGuardMix training set are included as preservation targets. Their outputs are generated by the original guard and used as the correct-target signal in both branches.
The combined 450 examples are split into 360 training and 90 validation samples using a stratified split (80/20) that maintains the shortcut-to-preserve ratio in both subsets.
For Qwen3Guard-8B, which exhibits lower baseline vulnerability, only 248 examples (199 train / 49 val) are available after shortcut-sensitivity filtering.

\paragraph{Cross-prefix contrast for GuardReasoner.}
For the two GuardReasoner variants, which generate chain-of-thought reasoning traces, we use cross-prefix contrast rather than baseline contrast.
For each sample, we identify one cue that preserves the correct harmful prediction and another that induces a shortcut-driven flip.
The corresponding model-generated reasoning traces and final predictions define the correct-target and shortcut-target pairs, respectively.
This design controls for cue presence while isolating the shortcut-induced change in the final harmfulness decision, avoiding conflation with format differences in the reasoning traces.

\paragraph{Optimizer and hyperparameters.}
All masks are optimized using AdamW with a learning rate of 0.01 and BF16 mixed precision.
The granularity is set to joint attention-head and MLP-neuron masking for all guards.
By default, each mask is optimized on selected examples from WildGuardMix for 30 epochs, with
\(\lambda_{\mathrm{corr}}=10\),
\(\lambda_{\mathrm{sc}}=1\), and
\(\gamma=10\).
Best masks are selected by the lowest validation loss (weighted sum of correct-target and shortcut-target cross-entropy, excluding sparsity) across all epochs.
The random seed is fixed at 42.

\paragraph{Permanent component suppression.}
After optimization, components assigned to the complementary branch (sigmoid gate value \(\leq 0.5\)) are permanently suppressed.
For attention heads, the corresponding slices of the output projection weight matrix are zeroed. For MLP neurons, the corresponding slices of the gate, up, and down projection weights are zeroed.
The resulting guard uses only the retained branch and requires neither a mask wrapper nor an additional inference pass, introducing zero overhead at deployment.

\paragraph{Computational cost.}
All masks are optimized on a single NVIDIA A100 80GB GPU without updating any model parameter, each converging within approximately a few hours.

\section{Ablation Study}
\label{app:ablation}

\paragraph{Component granularity.}
Table~\ref{tab:ablation}(a) compares joint, head-only, and neuron-only suppression for WG-7B.
Head-only suppression preserves clean harmfulness F1 but leaves head-position \(\dfr@3\) at 16.01\%.
Neuron-only suppression drives head-position \(\dfr@3\) to zero, but clean F1 drops from 75.85\% to 41.56\% and harmful recall falls to 30.14\%, indicating that strong suppression through neurons alone coincides with substantial damage to clean classification.
Joint suppression reduces head-position \(\dfr@3\) to 2.85\% while maintaining F1 at 76.32\% and raising harmful recall to 71.99\%, the most favorable overall trade-off.

\paragraph{Sparsity weight.}
Because attention heads account for a small fraction of the total maskable parameters, the overall masked percentage in the joint setting is dominated by the neuron mask ratio.
Table~\ref{tab:ablation}(b) varies the shared sparsity weight \(\gamma\).
Increasing \(\gamma\) from 5 to 15 shrinks the masked fraction from 10.4\% to 7.4\% at the cost of a modest rise in head-position \(\dfr@3\) from 2.49\% to 4.27\%, while clean harmfulness F1 remains stable across all three settings (76.32\% to 76.92\%).
We use \(\gamma=10\), which yields a head-position \(\dfr@3\) of 2.85\%.

\begin{table}[htb]
\centering
\small
\setlength{\tabcolsep}{1.0pt}
\begin{tabular}{lrrrrrcc}
\toprule
\multicolumn{8}{l}{\textit{(a) Component granularity}} \\
\cmidrule(lr){1-8}
Setting & Harm. F1\(\uparrow\) & Harm. R\(\uparrow\) & Head\(\downarrow\) & Middle\(\downarrow\) & Tail\(\downarrow\) & $M_{\mathrm{head}}$\% & $M_{\mathrm{neuron}}$\% \\
\midrule
No mask & 75.85 & 67.38 & 37.72 & 18.86 & 19.57 & -- & -- \\
\textbf{Joint} & 76.32 & 71.99 & 2.85 & 1.78 & 1.42 & 0.5 & 8.6 \\
Heads & 76.50 & 69.86 & 16.01 & 7.47 & 7.12 & 11.0 & 0 \\
Neurons & 41.56 & 30.14 & 0.00 & 0.71 & 3.91 & 0 & 15.4 \\
\midrule
\multicolumn{8}{l}{\textit{(b) Shared sparsity weight}} \\
\cmidrule(lr){1-8}
\(\gamma\) & Harm. F1\(\uparrow\) & Harm. R\(\uparrow\) & Head\(\downarrow\) & Middle\(\downarrow\) & Tail\(\downarrow\) & \multicolumn{2}{c}{Mask\%} \\
\midrule
No mask & 75.85 & 67.38 & 37.72 & 18.86 & 19.57 & \multicolumn{2}{c}{--} \\
5  & 76.46 & 71.99 & 2.49 & 2.14 & 2.49 & \multicolumn{2}{c}{10.4} \\
10 & 76.32 & 71.99 & 2.85 & 1.78 & 1.42 & \multicolumn{2}{c}{8.6} \\
15 & 76.92 & 70.92 & 4.27 & 2.85 & 3.20 & \multicolumn{2}{c}{7.4} \\
\bottomrule
\end{tabular}
\caption{Ablations for WG-7B on WildGuardTest.}
\label{tab:ablation}
\end{table}

\end{document}